\documentclass[letterpaper, 10 pt, conference]{ieeeconf}  

\IEEEoverridecommandlockouts                              

\usepackage[T1]{fontenc}
\usepackage{mathptmx}
\usepackage{multicol}
\usepackage{multirow}
\usepackage{tabularx}
\usepackage{float}
\usepackage{amsmath}
\usepackage{amssymb}
\usepackage{graphicx}
\usepackage{color}
\usepackage{caption}
\usepackage{stfloats}
\usepackage{booktabs}
\usepackage{cite}
\usepackage{hyperref}

\newcommand{\rev}[1]{#1}

\title{\LARGE \bf
HuMiT: Low-Latency Whole-Body Humanoid Teleoperation \\ via \rev{Compact} Reference Tracking
}

\newcommand{\authorrowbreak}{%
    \and[\hfill\mbox{}\par\vspace{0.8em}\mbox{}\hfill]%
}

\author{
    \authorblockN{Liu Cao}
    \authorblockA{Tsinghua IIIS}
    \and
    \authorblockN{Botian Xu}
    \authorblockA{The Chinese University of Hong Kong}
    \and
    \authorblockN{Ruoqu Chen}
    \authorblockA{Tsinghua IIIS}
    \and
    \authorblockN{Jiajun Liu}
    \authorblockA{Renmin University of China}
    \authorrowbreak
    \authorblockN{Mingzhi Pei}
    \authorblockA{Tsinghua College AI}
    \and
    \authorblockN{Zihao Wang}
    \authorblockA{Tsinghua Weiyang College}
    \and
    \authorblockN{Zhongyu Li}
    \authorblockA{The Chinese University of Hong Kong}
    \and
    \authorblockN{Mengdi Xu}
    \authorblockA{Tsinghua IIIS}
}

\begin{document}

\maketitle
\thispagestyle{empty}
\pagestyle{empty}

\begin{abstract}

Real-time whole-body humanoid teleoperation remains challenging due to the computational latency introduced by optimization-based retargeting pipelines and future-frame conditioning.
Existing motion tracking policies typically require dense reference targets including full joint positions, body-part transformations, and even future frames, which tightly couples the controller to expensive upstream modules.
In this work, we present \textit{HuMiT}, a whole-body teleoperation system built on a unified tracking policy that requires only a \emph{\rev{compact}} reference target, consisting of root height, root velocity, and \rev{the SE(3) poses of four sparse keypoints} at the current frame, yet achieves competitive tracking fidelity compared to methods relying on more diverse reference states including \rev{complete} joint positions, \rev{dense} body-part transformations, and even \rev{future frames}.
Experiments show that our policy achieves a communication-to-inference latency of $\sim$20\,ms measured between headset streaming and policy output and a video-based optical flow latency of \rev{65\,ms (median) across different motion types}.
We validate the system on a Unitree G1 humanoid, showcasing split-second reactive responses such as catching a tossed ball, as well as coordinated loco-manipulation capabilities. Project page: \url{https://xiaohu-art.github.io/Track/}.

\end{abstract}

\begin{figure*}[hb]
    \centering
    \includegraphics[width=\linewidth]{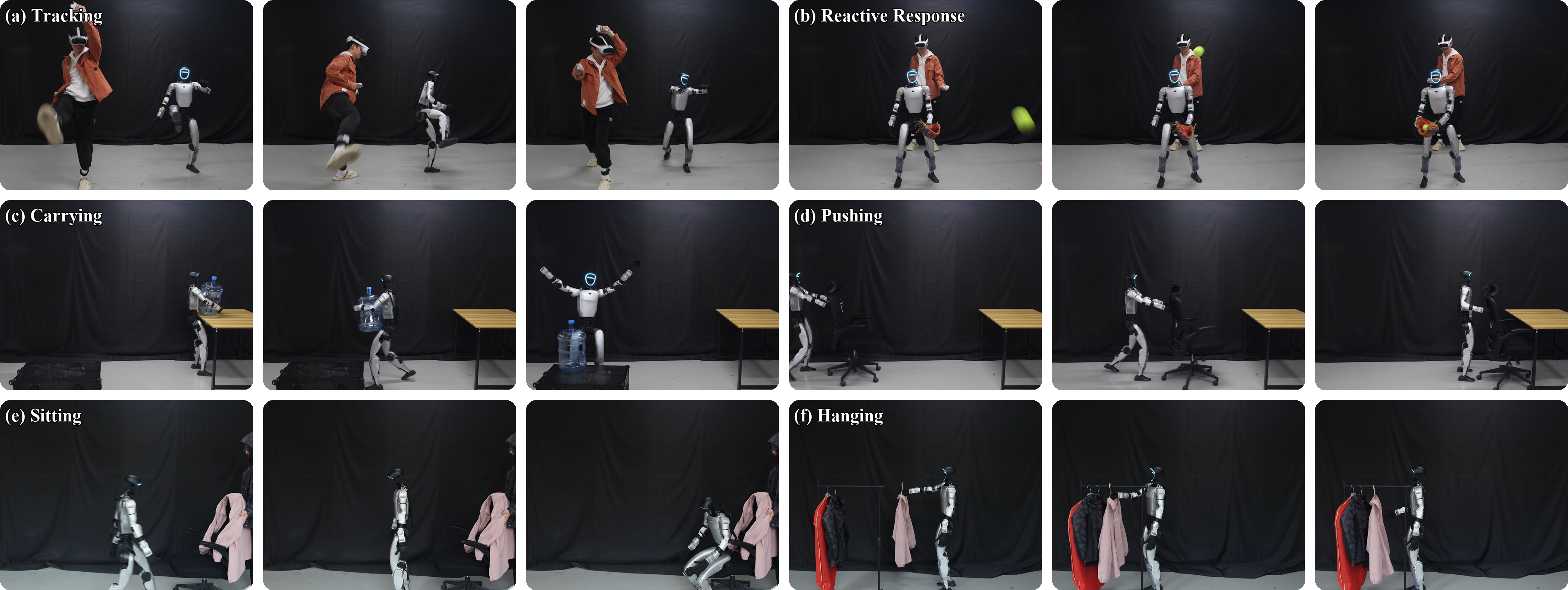}
    \caption{Real-world teleoperation with HuMiT. By removing optimization-based retargeting and future-frame conditions, HuMiT supports a wide range of dynamic whole-body skills with a \rev{compact}-reference tracking policy:
    \textbf{(a)} Accurate whole-body \emph{tracking} of operator motion, including front kick, side kick and boxing; 
    \textbf{(b)} \emph{reactive} ball catching enabled by low end-to-end latency;
    \textbf{(c)} \emph{carrying} a water jug across the workspace;
    \textbf{(d)} \emph{pushing} a wheeled chair while walking;
    \textbf{(e)} walking up to a chair, turning around and \emph{sitting} down;
    \textbf{(f)} reaching overhead to \emph{hang} a coat.}
    \label{fig:teaser}
\end{figure*}
\section{INTRODUCTION}

Humanoid robots hold unique promise for operating in human-centric environments due to their morphology that naturally matches the workspaces and tools designed for people.
Teleoperation is a practical paradigm for human-in-the-loop robot control.
It provides a scalable mechanism for collecting diverse demonstration data to train autonomous policies,
and also allows human operators to intervene in high-risk scenarios or situations where autonomous systems are likely to fail.

However, building an effective whole-body teleoperation system remains challenging. A primary issue is \textbf{large teleoperation latency}, which arises from two sources. First, optimization-based retargeting \cite{luo2023perpetual} introduces substantial per-frame delay and can suffer from numerical instability. Second, existing learning-based tracking policies rely heavily on \emph{full reference states} as inputs, including complete joint positions, body-part transformations, and their velocities \cite{liao2025beyondmimic, zeng2025behavior}.
Moreover, many of these methods condition the policy on \emph{future} reference frames to enable anticipatory control \cite{luo2025sonic, li2026omniclone}, which further exacerbates the latency problem.
A secondary challenge is \textbf{limited whole-body coordination}, as many approaches decouple upper- and lower-body control \cite{cheng2024expressive, ji2024exbody2, li2025amo}, which is particularly restrictive for dynamic tasks. 
In addition, hardware-centric solutions based on exoskeletons \cite{ben2025homie, fang2025airexo, yang2024ace} or marker-based motion capture \cite{ze2025twist, ze2025twist2} further constrain deployment flexibility.

In this work, we investigate a simple but consequential question: \emph{does high-fidelity whole-body motion tracking truly require a full reference state \rev{at deployment}?}
We find that a \textbf{\rev{compact reference target (CRT)}}, consisting only of a root height, root linear and angular velocity commands, and \rev{the SE(3) poses of four sparse keypoints (bilateral wrists and ankles)} at the current frame, is sufficient \rev{as the deployment-time input of} a unified whole-body tracking policy that achieves competitive tracking performance compared to methods relying on full reference states (i.e., complete joint positions, \rev{dense body-part states}, their velocities, and future frames).
This \rev{compact} reference design dramatically reduces the information that the upstream sensing module must provide, and it removes the retargeting pipeline from the online control loop entirely, resulting in significantly lower end-to-end latency.

We present \textbf{HuMiT}, a low-latency whole-body teleoperation system for humanoid robots.
HuMiT employs lightweight human motion sensing to produce \rev{compact} reference targets, which are then tracked by a whole-body policy trained in parallel simulation.
As shown in Fig.~\ref{fig:teaser}, HuMiT enables a wide range of dynamic whole-body skills that are difficult to achieve with high-latency or decoupled systems.

Our contributions are as follows:
\begin{itemize}
    \item \textbf{\rev{Compact} reference suffices \rev{at deployment}.}
    Through cross-method comparison and ablations under identical conditions, we show that a compact reference target \rev{matches full-reference policies}.
    \item \textbf{Low-latency teleoperation system.}
    By removing optimization-based retargeting and future-frame conditioning from the online loop, HuMiT achieves a communication-to-inference latency of $\sim$20\,ms measured between headset streaming and policy output and a video-based optical flow latency of \rev{65\,ms (median) between operator and robot motion}, while maintaining robust tracking performance.
    \item \textbf{Dynamic whole-body teleoperation for humanoids.}
    We validate HuMiT on a Unitree G1 humanoid executing reactive catching and coordinated loco-manipulation, confirming that the unified \rev{compact}-reference design supports tightly coupled whole-body control under low-latency constraints.
\end{itemize}

\section{Related Work}

\subsection{Learning-Based Humanoid Whole-Body Control}
\label{sec:2-A}

\textbf{Decoupled vs.\ unified control.}
Partitioning whole-body control into separate upper- and lower-body policies \cite{cheng2024expressive, ji2024exbody2, li2025amo} simplifies learning but fundamentally limits movements that couple limb motion and base stability, whereas unified policies integrate all degrees of freedom, enabling true loco-manipulation at the cost of increased training complexity.

\textbf{Reference target design.}
Among unified tracking controllers, prior methods differ in what reference information the policy observes.
BeyondMimic \cite{liao2025beyondmimic} conditions the policy on joint positions and joint velocities as reference targets.
BFM \cite{zeng2025behavior} augments this with body-part local positions, providing denser spatial supervision.
SONIC \cite{luo2025sonic} and OmniClone \cite{li2026omniclone} further extend the reference to include future frames for anticipatory control.
Despite these differences in reference richness, all of these methods share the assumption that dense reference targets are necessary for high-fidelity tracking, an assumption that remains largely unexamined.

\subsection{Humanoid Teleoperation Systems}
\label{sec:2-B}
Existing teleoperation systems \cite{darvish2023teleoperation} differ mainly in their sensing modality and in the resulting end-to-end latency.

\textbf{Hardware-centric sensing.}
Exoskeleton-based systems \cite{ben2025homie, fang2025airexo, yang2024ace} capture the operator's upper body while the lower body is driven by joysticks or pedals, so sensing leads to decoupled control. Optical marker-based motion capture \rev{suits} \cite{ze2025twist} are accurate but expensive, require extensive calibration, and restrict the operator to a constrained workspace.

\textbf{Vision-based pose estimation.}
Camera-based methods estimate human body pose using models such as SMPL \cite{cheng2024expressive, li2025genmo, shen2024world, wang2024tram}, offering greater flexibility and lower hardware cost, but the computational overhead of pose estimation followed by optimization-based retargeting makes the high-frequency, low-latency throughput required for responsive teleoperation hard to reach. HumanPlus \cite{fu2024humanplus} and OmniH2O \cite{he2024omnih2o} adopt this approach and inherit its latency limitations.

\textbf{Direct extremity control.}
A concurrent work, ExtremControl \cite{xiong2026extremcontrol}, operates directly on \rev{the SE(3) poses of six links (pelvis, torso, hands, and feet)} and incorporates velocity feedforward into the PD controller, reporting latency as low as 50\,ms with optical motion capture and 60\,ms with a VR-based setup.
\rev{Its deployable policy, however, is distilled from a privileged teacher that observes up to 32 future reference steps, and its reference is anchored by the operator's global pelvis and torso poses.
In contrast, HuMiT is trained in a single stage without distillation or any future-frame supervision, and replaces global pose anchoring with root height and velocity commands.}

\subsection{Motion Retargeting}
\label{sec:2-C}
Motion retargeting bridges the embodiment gap between the human operator and the target robot by converting captured human motion into robot-executable commands.
Classical approaches solve per-frame inverse kinematics (IK) or nonlinear optimization problems \cite{penco2018robust, luo2023perpetual} to map human joint positions onto the robot's kinematic structure.
While geometrically accurate, they introduce two practical drawbacks for teleoperation: (i)~the per-frame optimization incurs non-negligible latency that accumulates in the control loop, and (ii)~the numerical solvers are prone to instability (e.g., singularities, local minima), leading to jittery or infeasible reference commands.

In HuMiT, retargeting is performed \emph{offline} only during training data preparation. At deployment, the tracking policy directly consumes scaled human keypoints as reference targets, completely removing retargeting from the online loop and eliminating its associated latency and instability.

\section{Problem Formulation}
\label{sec:3}

We study low-latency whole-body humanoid teleoperation.
Let $\mathbf{S}^{h}_{t}$ denote the human kinematic state at time $t$, obtained from an upstream motion sensing device (e.g., VR headset with body tracking).
The state consists of root and body-part transformations in $SE(3)$:
$
\mathbf{S}^{h}_{t} = \{\mathbf{T}^{h}_{t, root},\; \mathbf{T}^{h}_{t,j}\}_{j \in K},
$
where $\mathbf{T}^{h}_{t, root} = (\mathbf{R}^{h}_{t, root}, \mathbf{x}^{h}_{t, root}) \in SE(3)$ is the root (pelvis) transformation expressed in the world frame, with $\mathbf{R}^{h}_{t, root} \in SO(3)$ denoting its orientation and $\mathbf{x}^{h}_{t, root} \in \mathbb{R}^3$ its position.
$\mathbf{T}^{h}_{t,j} = (\mathbf{R}^{h}_{t,j}, \mathbf{p}^{h}_{t,j}) \in SE(3)$ is the transformation of the $j$-th body part in the root frame, with $\mathbf{R}^{h}_{t,j} \in SO(3)$ its orientation and $\mathbf{p}^{h}_{t,j} \in \mathbb{R}^3$ its position.
$K$ denotes a predefined set of keypoints. In our setup, $|K|=12$, \rev{including the bilateral hips, knees, ankles, shoulders, elbows, and wrists}.

A teleoperation system must convert the human state stream $\mathbf{S}^{h}_{1:T}$ into robot-executable actions at a high frequency.
We decompose this into two stages: \textbf{reference generation}, in which a mapping $\mathcal{F}$ converts the human state into a robot-frame reference target $\mathbf{g}_{t} = \mathcal{F}(\mathbf{S}^{h}_{t})$, and \textbf{motion tracking}, in which a policy $\pi$ maps onboard observations and the reference target to joint-space actions $\mathbf{a}_t \sim \pi(\cdot \mid \mathbf{o}_t,\, \mathbf{g}_{t})$.
The key design question we investigate is \emph{how compact $\mathbf{g}_{t}$ can be} while still enabling high-fidelity whole-body tracking.

\section{Method}
\label{sec:4}

\begin{figure*}[t]
  \centering
  \includegraphics[width=\linewidth]{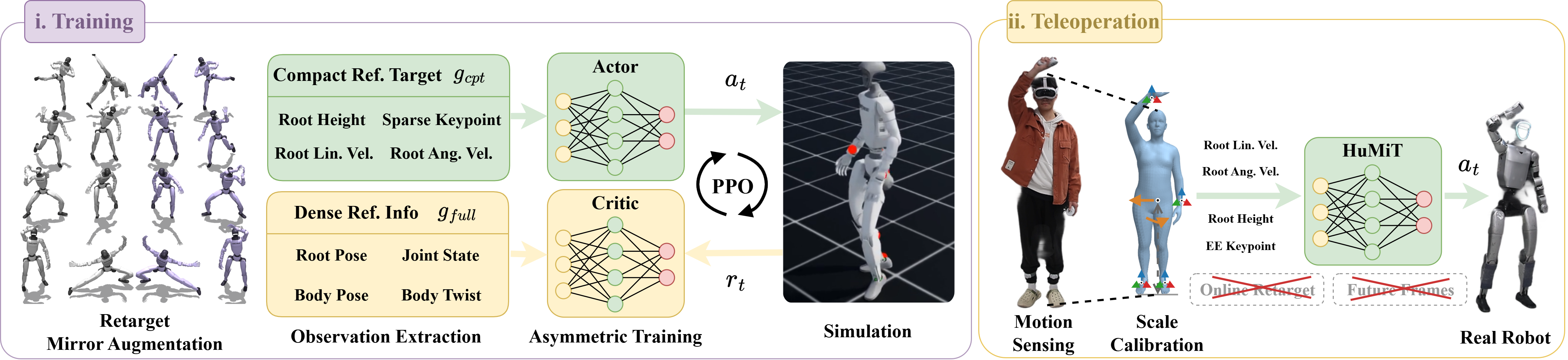}
  \caption{Method overview. 
  Human motion passes through scale calibration to produce a \rev{compact} reference target (\rev{$g_{cpt}$}). Stages required by traditional pipelines (inverse kinematics, future-frame buffering) are eliminated.
  During training, the Actor $\pi$ receives only the \rev{compact} reference \rev{$g_{cpt}$} and proprioception $s_{prop}$, and outputs actions $a_t$.
  The Critic additionally observes privileged states $s_{priv}$ and full retargeted references $g_{full}$, along with dense reward $r_t$, enabling accurate value estimation.
  Both networks are updated jointly via PPO.}
  \label{fig:system}
  \vspace{-10pt}
\end{figure*}

\subsection{System Overview}
Fig.~\ref{fig:system} illustrates the HuMiT pipeline.
Human motion is captured by a VR headset and converted into a \rev{compact reference target (CRT)} \rev{$g_{cpt}$} through a one-time scale calibration.
The \rev{CRT} is then consumed by a reinforcement-learning-based \emph{whole-body tracking policy} $\pi$ that outputs joint-space commands for the Unitree G1 humanoid ($N=29$ DoF).

The central design choice in HuMiT is that $\pi$ observes only the \rev{CRT}, consisting of root height $h^{r}_t$, root linear velocity $\mathbf{v}^{r}_t$, root angular velocity $\boldsymbol{\omega}^{r}_t$, and \rev{the root-relative SE(3) poses $(\mathbf{R}^{r}_{t,j}, \mathbf{p}^{r}_{t,j})$ of a sparse keypoint subset $j \in K_c \subset K$ (bilateral wrists and ankles, $|K_c|=4$), without requiring joint-level references $\mathbf{q}^{r}_t$ and $\mathbf{\dot{q}}^{r}_t$, the states of the remaining body parts, or any future-frame information}.

This reduces the reference generation stage in Section~\ref{sec:3} to two lightweight components:
(1)~a \emph{motion sensing} module that captures the operator's body keypoints via VR headset (PICO 4 Ultra) with motion trackers, and
(2)~a \emph{scale calibration} module that maps human keypoint positions to the robot's body positions in closed form, in place of an optimization-based retargeting loop.

\subsection{\texorpdfstring{From Human Motion to \rev{Compact} Reference Target}{From Human Motion to Compact Reference Target}}
\label{sec:4-A}
 
\textbf{Motion sensing.}
\rev{Our sensing front-end is entirely consumer-grade as in Fig.~\ref{fig:system}: a PICO 4 Ultra VR headset, two handheld PICO controllers, and two PICO Motion Trackers strapped to the operator's lower legs.
From this setup, the PICO body-tracking runtime infers a full-body skeleton of 24 points with 6 DoF each at approximately 80\,Hz and streams it to the control host over Wi-Fi. From this stream we extract the SE(3) poses of the four keypoints in $K_c \subset K$.
Putting on the entire wearable set takes less than a minute, and the system imposes no workspace constraint beyond Wi-Fi coverage, in contrast to exoskeleton or optical motion-capture pipelines.}
 
\textbf{Scale calibration.}
To bridge the morphological difference between the human operator and the robot, we perform a one-time calibration following GMR \cite{araujo2025retargeting}.
At the start of each teleoperation session, the operator assumes a reference pose (e.g., T-pose).
We compute per-limb scale factors $\alpha_j$ by comparing the operator's limb segment lengths to those of the humanoid robot, and apply these factors to map incoming human keypoint positions into the robot's body frame\rev{, while keypoint orientations require no morphological rescaling and are passed through unchanged}:
\begin{equation*}
\mathbf{p}^{r}_{t,j} = \alpha_j \cdot \mathbf{p}^{h}_{t,j}, \qquad
\rev{\mathbf{R}^{r}_{t,j} = \mathbf{R}^{h}_{t,j}}, \qquad j \in \rev{K_c}.
\end{equation*}
The root height $h^{r}_{t}$ is obtained as the $z$-component of the scaled root position, and the root velocity is computed via finite differences of the root transformation.
All mappings are closed-form and add negligible latency ($<$1\,ms).

\subsection{Whole-Body Motion Tracking Policy}
\label{sec:4-B}
 
We train a unified whole-body tracking policy $\pi$ via reinforcement learning in simulation.
Training proceeds in two stages: (1)~offline data preparation, in which human motion clips are retargeted to the robot via IK to produce full reference states $g_{full}$, 
and (2)~online RL training, in which $\pi$ learns to track the \rev{CRT} using PPO with an asymmetric actor--critic architecture.
The full reference $g_{full}$ is used only to (i)~initialize feasible simulator states, (ii)~compute dense reward signals, and (iii)~serve as an additional critic input.
It is worth noting that the actor $\pi$ itself never observes $g_{full}$.

\textbf{Training data.}
We train on motion sequences from AMASS \cite{mahmood2019amass}, LAFAN1 \cite{harvey2020robust, lvhaidong_lafan1_retargeting_dataset}\rev{, and BONES-SEED~\cite{luo2025sonic}} after filtering out physically infeasible clips.
\rev{Every clip is further augmented with its sagittal-plane mirror, which swaps the left and right joints and reflects the root trajectory about the vertical plane through the initial heading.}
For each human clip, we solve an optimization-based inverse kinematics problem to obtain a robot-frame reference trajectory
\begin{equation*}
\{g_{full}\}^{T}_{t=1} = 
\{\mathbf{x}^{r}_{t, root},\; 
\mathbf{R}^{r}_{t, root},\; 
\mathbf{v}^r_{t},\;
\boldsymbol{\omega}^r_{t},\;
\mathbf{p}^{r}_{t,j},\; 
\mathbf{R}^{r}_{t,j},\; 
\mathbf{q}^{r}_{t},\;
\mathbf{\dot{q}}^r_{t}\}_{t=1}^{T},
\label{eq:sfull}
\end{equation*}
which contains both link-level and joint-level references.
We emphasize that this offline retargeting is performed \emph{only once} during data preparation and is \emph{never} part of the online teleoperation loop.
 
\textbf{Observations.}
We employ an asymmetric actor--critic scheme.
Symbols without the $(\cdot)^{r}$ superscript refer to the robot's \emph{measured} state as opposed to the \emph{reference} $(\cdot)^{r}$.

The observations are partitioned into four groups:
\begin{align*}
s_{prop} &= [\boldsymbol{\omega}_{t},\; \boldsymbol{\phi}_t,\; \mathbf{q}_t,\; \dot{\mathbf{q}}_t,\; \mathbf{a}_{t-1}], \\
s_{priv} &= [\mathbf{v}_{t},\; \mathbf{p}_{t, j},\; \mathbf{R}_{t, j}], \\
g_{full} &= [\mathbf{x}^{r}_{t,root},\; 
\mathbf{R}^{r}_{t,root},\; \mathbf{v}^r_{t},\; 
\boldsymbol{\omega}^r_{t},\; \mathbf{p}^{r}_{t,j},\; 
\mathbf{R}^{r}_{t,j},\; 
\mathbf{q}^{r}_{t},\;
\mathbf{\dot{q}}^r_{t}], \; \\
\rev{g_{cpt}} &= [h^{r}_{t},\; \mathbf{v}^{r}_{t},\; \boldsymbol{\omega}^{r}_{t},\; \rev{\{\mathbf{p}^{r}_{t,j},\, \mathbf{R}^{r}_{t,j}\}_{j \in K_c}}] \subset g_{full},
\end{align*}
where $s_{prop}$ is the robot proprioceptive state ($\boldsymbol{\omega}_t$ is the measured root angular velocity \rev{in the root frame}, $\boldsymbol{\phi}_t$ the gravity vector \rev{projected into the root frame}, $\mathbf{q}_t, \dot{\mathbf{q}}_t$ the measured joint positions and velocities, and $\mathbf{a}_{t-1}$ the previous action).
The privileged group $s_{priv}$ contains ground-truth body states available only in simulation \rev{(the root linear velocity $\mathbf{v}_{t}$ in the root frame, and the root-relative body poses $\mathbf{p}_{t,j}$, $\mathbf{R}_{t,j}$)}.
The full reference $g_{full}$ is defined in Eq.~\eqref{eq:sfull}, and \rev{$g_{cpt}$} is a strict subset of $g_{full}$ that retains only the root height \rev{(the world-frame $z$-coordinate of the root above the ground)}, root velocities \rev{(in the root frame)}, and \rev{the root-relative sparse keypoint poses over $K_c$ (orientations encoded in the 6D rotation representation)}, omitting joint-level targets, \rev{the dense states of the remaining body parts}, and future frames.

The actor and critic observations are defined as:
\begin{equation*}
o_\pi := [s_{prop},\; \rev{g_{cpt}}], \qquad
o_v := [s_{prop},\; s_{priv},\; g_{full}].
\end{equation*}
 
\textbf{Action and low-level control.}
The policy outputs a joint-space offset $\mathbf{a}_t \in \mathbb{R}^{N}$ that yields the joint targets $\mathbf{q}_{target} = \mathbf{q}_{0} + \sigma\,\mathbf{a}_t$, where $\mathbf{q}_{0}$ is the nominal standing configuration and $\sigma$ scales the output to the feasible joint range. A PD controller then computes motor torques $\boldsymbol{\tau} = K_p(\mathbf{q}_{target} - \mathbf{q}_t) - K_d\,\dot{\mathbf{q}}_t$.

\renewcommand{\arraystretch}{1.3}
\begin{table}[h]
\caption{Reward Terms. Rotations are compared using the axis--angle residual ($\ominus$).}
\centering
\label{tab:reward_terms}
\begin{tabular}{l c c}
\toprule
\textbf{Reward Term} & \textbf{Error Term} & \textbf{Weight} \\
\midrule
Anchor position          & $\mathbf{x}^{r}_{t, root} - \mathbf{x}_{t, root}$              & $1.0$ \\
Anchor orientation       & $\mathbf{R}^{r}_{t,root} \ominus \mathbf{R}_{t,root}$         & $1.0$ \\
Body position            & $\mathbf{p}^{r}_{t,j} - \mathbf{p}_{t,j}$                                    & $2.0$ \\
Body orientation         & $\mathbf{R}^{r}_{t,j} \ominus \mathbf{R}_{t,j}$                              & $2.0$ \\
Body linear velocity     & $\dot{\mathbf{p}}^{r}_{t,j} - \dot{\mathbf{p}}_{t,j}$                        & $2.0$ \\
Body angular velocity    & $\boldsymbol{\omega}^{r}_{t,j} - \boldsymbol{\omega}_{t,j}$                  & $2.0$ \\
Feet slip (penalty)      & $\sum_{j\in \text{feet}} \mathbf{v}_{t,j} \cdot \mathbb{I}_{contact}$ & $-1.0$ \\
Action rate (reg.)       & $\mathbf{a}_{t} - \mathbf{a}_{t-1}$                                          & $-0.2$ \\
\bottomrule
\end{tabular}
\vspace{-15pt}
\end{table}
\renewcommand{\arraystretch}{1.0}

\textbf{Reward.}
The tracking reward combines exponential tracking terms with regularization penalties:
$
r_{\text{track}} = \sum_i w_i \exp\!\left(-\frac{e_i}{\delta_i^2}\right),
$
where $e_i$ is the $i$-th tracking error and $w_i$ is its weight.
Table~\ref{tab:reward_terms} lists all reward terms, where the ``anchor'' terms refer to the robot root (pelvis).
Notably, the reward function uses the full retargeted reference (body positions, body orientations, velocities) for dense error signals during training, even though $\pi$ itself only observes the \rev{compact} reference.
This asymmetry between the reward supervision and the policy observation is key: it provides rich learning gradients while keeping the deployment interface compact.

\textbf{Early termination and domain randomizations.}
We terminate episodes when the root translation deviation exceeds 0.4\,m,
root orientation deviation exceeds 0.8, or keypoint deviation exceeds 0.4\,m.
To improve sim-to-real transfer, we randomize external pushes, contact friction and restitution, link masses (with stronger perturbations on end-effectors to approximate small payload variations), motor gains, armature parameters, and initial joint states.
 
\textbf{Training and deployment.}
We train with PPO \cite{schulman2017proximal} across 8192 parallel environments in Isaac Sim \cite{mittal2025isaac}.
For real-world deployment on the Unitree G1, the trained Actor policy receives proprioceptive feedback from onboard sensors and the \rev{CRT} streamed from the PICO headset, with no additional processing required.

\section{Results}
\label{sec:5}

We design experiments to investigate three questions:

\begin{itemize}

\item \textbf{(Q1) System latency:} What is the latency of HuMiT between human operator and robot motion, and how does it compare to state-of-the-art systems that rely on optimization-based retargeting and future frames?
\item  \textbf{(Q2) Sufficiency of \rev{compact} reference:} Does the \rev{CRT} suffice \rev{as the policy input at deployment}, compared to policies that observe full reference states?
\item \textbf{(Q3) Real-world teleoperation:} Can HuMiT achieve dynamic whole-body \rev{teleoperation} in the real world, including tasks that require coordinated loco-manipulation and fast reactive response?
    
\end{itemize}

\subsection{End-to-End Latency (\textbf{Q1})}
\label{sec:5-A}

\rev{Conditioning a policy on $k$ future reference frames delays at least $k\,\Delta t$ of reference buffering (${\geq}200$\,ms
for 10 frames at 20\,ms spacing).}

\begin{figure}[!t]
  \centering
  \includegraphics[width=\linewidth]{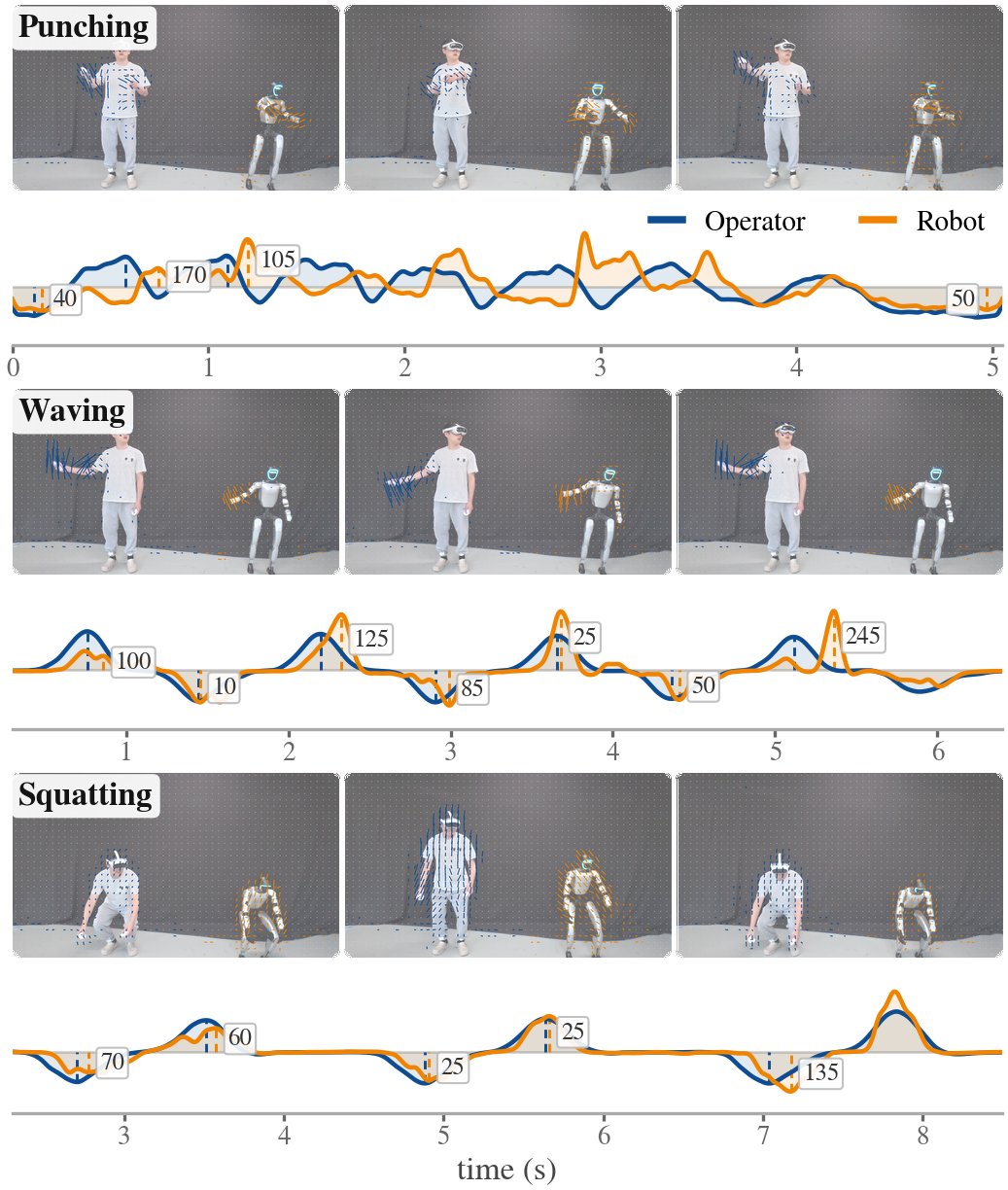}
  \caption{\rev{End-to-end latency across motion types. For each motion, the flow-overlay frames of the operator (blue) and the robot (orange) are shown above. White boxes give per-pair delays in ms.}}
  \label{fig:latency}
  \vspace{-15pt}
\end{figure}

\rev{
\textbf{Internal timestamped measurement.} We record timestamps from
the arrival of a skeleton packet streamed by the headset to the policy's
action, covering wireless transmission, CRT assembly, and policy
inference. This yields a communication-to-inference latency of
${\sim}20$\,ms, excluding the sensing latency and the
robot's actuation response.

\textbf{External optical-flow estimation.} To capture the full
operator-to-robot delay including physical execution, we record the operator
and the robot in the same camera frame, compute dense optical flow between
consecutive frames within an operator region and a robot region, and reduce
each region to one standardized motion signal by averaging its flow along the
vertical direction.
We then detect prominent peaks in the operator signal and match each to
the robot extremum of the same sign, and the per-peak latency is the time offset
of the matched pair (Fig.~\ref{fig:latency}).
We record three motion types at 200\,fps (5\,ms temporal resolution):
punching, squatting, and waving, yielding median (mean) latencies of
78 (91)\,ms with $n{=}4$ matched pairs, 60 (63)\,ms with $n{=}5$, and
85 (91)\,ms with $n{=}7$, respectively. The pooled median (mean) over all
16 pairs is 65 (83)\,ms.

\textbf{Baseline comparisons.} Table~\ref{tab:latency} applies the same
protocol to the systems we can deploy with our sensing front-end.
Shortening the lookahead of SONIC from 10 to 4 future frames (SONIC-LL) drops
the delay from 245 to 145\,ms, in line with the 120\,ms of buffering removed.
TWIST2 buffers nothing but retargets online and measures 120\,ms, while
HuMiT, which does neither, measures 65\,ms.}

\begin{table}[h]
\centering
\caption{\rev{End-to-end teleoperation latency, measured with the external
optical-flow protocol.}}
\label{tab:latency}
\setlength{\tabcolsep}{3pt}
\renewcommand{\arraystretch}{1.15}
\footnotesize
\begin{tabular*}{\linewidth}{@{\extracolsep{\fill}}l c c c@{}}
\toprule
\textbf{Method} & \textbf{Buffering (ms)} & \textbf{Online retarget.} & \textbf{End-to-end (ms)} \\
\midrule
SONIC~\cite{luo2025sonic} & 200 & $\times$      & 245 \\
SONIC-LL                  &  80 & $\times$      & 145 \\
TWIST2~\cite{ze2025twist2}&   0 & $\checkmark$  & 120 \\
\midrule
HuMiT (ours)              & \textbf{0} & $\times$ & \textbf{65} \\
\bottomrule
\end{tabular*}
\renewcommand{\arraystretch}{1.0}
\vspace{-15pt}
\end{table}

\subsection{\texorpdfstring{Sufficiency of \rev{Compact} Reference Targets (\textbf{Q2})}{Sufficiency of Compact Reference Targets (Q2)}}
\label{sec:5-B}

We evaluate whether the \rev{CRT} is sufficient \rev{as the policy input at deployment} through two complementary comparisons:
(1)~against existing methods that use richer reference observations under different training frameworks, and
(2)~an ablation within HuMiT itself that isolates the effect of reference richness.

\subsubsection{Metrics}
\rev{We report five complementary metrics over $T$ evaluated frames, the keypoint set $K$ ($|K|=12$), and the $N=29$ actuated joints:
\begin{equation*}
E_{kp} = \frac{1}{T|K|}\sum_{t,j}\bigl\|\mathbf{x}^{r}_{t,j}-\mathbf{x}_{t,j}\bigr\|_{2},
\quad
E_{jpe} = \frac{1}{TN}\sum_{t}\bigl\|\mathbf{q}^{r}_{t}-\mathbf{q}_{t}\bigr\|_{1},
\end{equation*}
where $\mathbf{x}_{t,j}$ is the world-frame position of keypoint $j$, in contrast to the root-relative $\mathbf{p}_{t,j}$, so that the mean per-keypoint error $E_{kp}$ (mm) measures whole-body geometric accuracy and the mean joint position error $E_{jpe}$ (rad) joint-level accuracy.
The root velocity errors $E_{lin} = \frac{1}{T}\sum_{t}\|\mathbf{v}^{r}_{t}-\mathbf{v}_{t}\|_{2}$ (m/s) and $E_{ang} = \frac{1}{T}\sum_{t}\|\boldsymbol{\omega}^{r}_{t}-\boldsymbol{\omega}_{t}\|_{2}$ (rad/s) capture root tracking fidelity, and the success rate (SR, \%) is the fraction of rollouts that reach the end of the clip without triggering the early-termination criteria of Section~\ref{sec:4-B}.
}

\subsubsection{Baselines}
\rev{We compare against five external baselines spanning a spectrum of reference richness, plus our controlled full-reference variant.
\textbf{SONIC}~\cite{luo2025sonic} is a unified tracking policy trained on BONES-SEED, conditioned on dense reference states over \emph{10 future frames} (${\sim}200$\,ms of lookahead), and \textbf{SONIC-LL} is its low-latency version, which shortens the reference lookahead to \emph{4 future frames} (${\sim}80$\,ms) for more responsive teleoperation.
\textbf{ScaleBFM (M/XL)}~\cite{zeng2026scaling} are behavior foundation models built on~\cite{zeng2025behavior} that track masked root-relative target poses of body links over \emph{6 future frames} ($\sim 120$ ms). We evaluate both officially released Humanoid Transformer checkpoints (M: 3.0M, XL: 9.9M parameters).
\textbf{ExtremControl}~\cite{xiong2026extremcontrol} is a concurrent low-latency method tracking the SE(3) poses of six links (pelvis, torso, hands, and feet), obtained via privileged teacher--student distillation.
}

To isolate the effect of reference richness from differences in architecture and training, we additionally train a \textbf{full-reference} variant of HuMiT (\textbf{HuMiT-Full}) in which the policy $\pi$ observes $g_{full}$ instead of \rev{$g_{cpt}$} in addition to proprioception $s_{prop}$.
All other components, including network architecture, reward function, domain randomization, and training procedure, remain identical\rev{, making HuMiT vs.\ HuMiT-Full the controlled comparison that isolates reference richness with the training methodology held fixed}.

\begin{table*}[!t]
\centering
\caption{\rev{Whole-body motion tracking evaluation. $E_{kp}$ (mm), $E_{jpe}$ (rad), $E_{lin}$ (m/s), $E_{ang}$ (rad/s), SR (\%). A compact reference target, without joint-level targets, dense per-link states, or future frames, is sufficient as the deployment-time input of a whole-body tracking policy with competitive fidelity.}}
\label{tab:motion_tracking}
\setlength{\tabcolsep}{2.6pt}
\renewcommand{\arraystretch}{1.15}
\footnotesize
\begin{tabular}{@{}l ccccc ccccc ccccc@{}}
\toprule
 & \multicolumn{5}{c}{\textbf{AMASS}} & \multicolumn{5}{c}{\textbf{LAFAN1}} & \multicolumn{5}{c}{\textbf{BONES-SEED}} \\
\cmidrule(lr){2-6} \cmidrule(lr){7-11} \cmidrule(l){12-16}
\textbf{Method}
 & $E_{kp}\!\downarrow$ & $E_{jpe}\!\downarrow$ & $E_{lin}\!\downarrow$ & $E_{ang}\!\downarrow$ & SR$\uparrow$
 & $E_{kp}\!\downarrow$ & $E_{jpe}\!\downarrow$ & $E_{lin}\!\downarrow$ & $E_{ang}\!\downarrow$ & SR$\uparrow$
 & $E_{kp}\!\downarrow$ & $E_{jpe}\!\downarrow$ & $E_{lin}\!\downarrow$ & $E_{ang}\!\downarrow$ & SR$\uparrow$ \\
\midrule
SONIC~\cite{luo2025sonic}
 & 316.0 & 0.116 & 0.141 & \textbf{0.312} & 79.8
 & 1492.6 & 0.097 & \textbf{0.221} & \textbf{0.394} & 20.0
 & 143.1 & 0.098 & 0.112 & 0.295 & 98.2 \\
SONIC-LL
 & 400.5 & \textbf{0.113} & 0.177 & 0.397 & 59.5
 & 1524.5 & 0.110 & 0.245 & 0.537 & 15.0
 & 229.9 & 0.098 & 0.136 & 0.384 & 96.5 \\
ScaleBFM-M~\cite{zeng2026scaling}
 & 73.2 & 0.123 & 0.145 & 0.444 & 92.6
 & \textbf{82.1} & 0.102 & 0.226 & 0.699 & 27.5
 & 60.2 & \textbf{0.079} & 0.136 & 0.435 & 97.7 \\
ScaleBFM-XL~\cite{zeng2026scaling}
 & \textbf{71.8} & 0.125 & \textbf{0.137} & 0.407 & \textbf{94.0}
 & 82.8 & 0.097 & 0.224 & 0.621 & 32.5
 & \textbf{58.5} & 0.081 & 0.128 & 0.403 & \textbf{98.7} \\
ExtremControl~\cite{xiong2026extremcontrol}
 & 329.5 & 0.246 & 0.259 & 0.675 & 60.1
 & 191.0 & 0.203 & 0.244 & 0.781 & 0.0
 & 220.2 & 0.138 & 0.203 & 0.638 & 66.3 \\
\midrule
HuMiT-Full (ours)
 & 307.9 & 0.132 & 0.146 & 0.399 & 87.5
 & 132.1 & 0.054 & 0.370 & 0.732 & \textbf{62.5}
 & 154.6 & 0.092 & \textbf{0.089} & 0.212 & 96.6 \\
HuMiT (ours)
 & 313.5 & 0.133 & 0.145 & 0.391 & 87.6
 & 148.0 & \textbf{0.035} & 0.370 & 0.716 & 52.5
 & 155.0 & 0.126 & 0.090 & \textbf{0.203} & 96.2 \\
\bottomrule
\end{tabular}
\renewcommand{\arraystretch}{1.0}
\vspace{-10pt}
\end{table*}

\subsubsection{Quantitative Results}

\rev{Table~\ref{tab:motion_tracking} shows that HuMiT attains competitive tracking fidelity across all three datasets while observing the least reference information among the compared methods and no future frames.

\textbf{Baseline comparisons.}
On AMASS, HuMiT tracks at $E_{kp}=313.5$\,mm with an $87.6\%$ success rate, and its root-velocity errors ($E_{lin}=0.145$\,m/s, $E_{ang}=0.391$\,rad/s) are on par with the strongest baselines.
On LAFAN1, the most dynamic dataset, it attains the highest success rate among all external baselines ($52.5\%$) and the lowest joint error of all compared methods ($E_{jpe}=0.035$\,rad).
On BONES-SEED, it reaches a $96.2\%$ success rate with the lowest root angular-velocity error of all compared methods ($E_{ang}=0.203$\,rad/s) and an $E_{lin}$ of $0.090$\,m/s within $0.001$\,m/s of the best.
The world-frame $E_{kp}$, however, favors ScaleBFM, which observes global root-position targets and corrects drift online.

\textbf{Ablation: compact vs.\ full reference.}
Under identical training conditions, HuMiT and HuMiT-Full are nearly indistinguishable on AMASS and BONES-SEED.
The most notable gap is on LAFAN1, indicating that a dense reference target still helps for tracking highly dynamic motion.}

\begin{table}[!t]
\centering
\caption{\rev{Real-world teleoperation reported as
successful trials over total trials with a single operator. SONIC-LL and TWIST2 never catch the ball (0/10).}}
\label{tab:success}
\setlength{\tabcolsep}{2pt}
\renewcommand{\arraystretch}{1.15}
\scriptsize
\begin{tabular*}{\linewidth}{@{\extracolsep{\fill}}l c l@{}}
\toprule
\textbf{Task} & \textbf{Succ./trials} & \textbf{Dominant failure mode} \\
\midrule
Reactive catching & 2 / 10 & Arm workspace differs from operator's \\
Carrying a water jug & 9 / 10 & Jug tips out of the grasp \\
Pushing a wheeled chair & 5 / 10 & Chair heading hard to steer \\
Sitting on a chair      & 7 / 10 & Chair out of view after turning \\
Hanging a coat          & 6 / 10 & Hanger orientation uncontrolled \\
\bottomrule
\end{tabular*}
\renewcommand{\arraystretch}{1.0}
\vspace{-15pt}
\end{table}

\subsection{Real-World Teleoperation (\textbf{Q3})}
\label{sec:5-C}

We deploy HuMiT on a Unitree G1 to evaluate three capabilities that are directly enabled by its low-latency, unified-policy design: accurate real-time tracking, fast reactive responses, and coordinated loco-manipulation. Snapshots of all real-world demonstrations are shown in Fig.~\ref{fig:teaser}\rev{, and Table~\ref{tab:success} reports the outcome of ten trials per task together with the dominant failure mode we observed}.

\textbf{Real-time tracking.} We first stream live operator motion to the robot and evaluate qualitatively how closely the G1 mirrors the human (Fig.~\ref{fig:teaser}a). The robot reproduces a wide range of whole-body poses driven by the \rev{compact} reference target alone.
 
\textbf{Reactive catching.} To validate the low-latency advantage, we conduct a task as shown in Fig.~\ref{fig:teaser}b:
A ball is thrown toward the robot, which mirrors the operator's motion in real time and catches the ball with its glove.
This task is inherently difficult for high-latency systems, since the robot has to react within the flight time of the ball.
\rev{HuMiT catches the ball in 2 of 10 trials, whereas SONIC-LL and TWIST2 never succeed, and its remaining failures come from the mismatch between the operator's and the robot's arm workspace rather than from a delayed response.}

\textbf{Coordinated loco-manipulation.} We further evaluate four loco-manipulation tasks that require tight coordination between locomotion and manipulation (Fig.~\ref{fig:teaser}c--f). In all of them, the unified policy maintains balance during locomotion without degrading manipulation accuracy.
\rev{SONIC-LL and TWIST2 remain stable on these four tasks, but their higher latency requires the operator to pause and correct after each motion, which makes teleoperated data collection substantially slower.} These behaviors are difficult to obtain with decoupled upper-/lower-body controllers, where the lower body is typically driven by joystick or pedal commands that cannot be tightly synchronized with arm motions.

\section{CONCLUSIONS AND LIMITATIONS}
\label{sec:conclusion}

We presented HuMiT, showing that a \rev{compact} reference target is sufficient \rev{as the deployment-time input of} a whole-body tracking policy with competitive fidelity, achieving a communication-to-inference latency of $\sim$20\,ms and an externally measured end-to-end latency of \rev{65\,ms (median)}.
Real-world experiments demonstrate that this latency enables reactive tasks such as catching a tossed ball, as well as coordinated loco-manipulation, validating the approach.

HuMiT also has several limitations.
First, the training data is dominated by upright locomotion and upper-body gestures, so motions involving ground contact such as kneeling or crawling are underrepresented, leading to degraded tracking accuracy or failure on these behaviors.
Second, current hardware and teleoperation policy design do not support high-payload loco-manipulation (e.g., carrying heavy objects while walking), as the policy has not been trained under such loading conditions.
Addressing these limitations through broader training distributions and load-aware policy learning is a promising direction for future work.





\bibliographystyle{IEEEtran}
\bibliography{references}

\end{document}